\documentclass[letterpaper, 10 pt, conference]{ieeeconf}  % Comment this line out if you need a4paper

\IEEEoverridecommandlockouts                              % This command is only needed if 
\usepackage{cite}
\usepackage{amsmath,amssymb,amsfonts}
\usepackage{algorithmic}
\usepackage{graphicx}
\usepackage{textcomp}
\usepackage{xcolor}

\usepackage{makeidx}

\usepackage{pgfplots}
\pgfplotsset{compat=newest}
\usepackage{multirow}
\usepackage{tabularx}
\usepackage{threeparttable}
\usepackage{booktabs}

\usepackage[draft,inline,nomargin]{fixme}
\fxusetheme{color}

\usepackage{array}
\usepackage{float}
\usepackage{graphbox}
\usepackage{subcaption}
\usepackage{tikz,tikz-3dplot}
\usetikzlibrary{fit,arrows,arrows.meta,automata,backgrounds,calc,chains,%
decorations.markings,decorations.pathreplacing,decorations.pathmorphing,%
matrix,positioning,shapes,shapes.geometric,shapes.symbols,spy,trees,tikzmark}
\usepackage{balance}
\usepackage[binary-units=true,product-units=single,per-mode=symbol,range-units=single,range-phrase=\,--\,,detect-all]{siunitx}
\DeclareSIUnit\pixel{px}
\usepackage{bm}
\usepackage{acronym}
\usepackage{tablefootnote}
\usepackage{hyperref}

\usepackage{caption}
\let\labelindent\relax
\usepackage[inline]{enumitem}

\usepackage{adjustbox}
\newcolumntype{R}[2]{%
    >{\adjustbox{angle=#1,lap=\width-(#2)}\bgroup}%
    l%
    <{\egroup}%
}
\newcolumntype{L}[1]{>{\raggedright\let\newline\\\arraybackslash\hspace{0pt}}m{#1}}

\title{\LARGE \bf
OC-SOP: Enhancing Vision-Based 3D Semantic Occupancy Prediction by Object-Centric Awareness
}

\author{Helin Cao and Sven Behnke% <-this % stops a space
\thanks{
	This research has been supported by MBZIRC prize money. All authors are with the Autonomous Intelligent Systems group, Computer Science Institute VI – Intelligent Systems and Robotics – and the Center for Robotics and the Lamarr Institute for Machine Learning and Artificial Intelligence, University of Bonn, Germany; {\tt\small caoh@ais.uni-bonn.de}}%
}

\begin{document}

\maketitle
\thispagestyle{empty}
\pagestyle{empty}

%%%%%%%%%%%%%%%%%%%%%%%%%%%%%%%%%%%%%%%%%%%%%%%%%%%%%%%%%%%%%%%%%%%%%%%%%%%%%%%%
\begin{abstract}
Autonomous driving perception faces significant challenges due to occlusions and incomplete scene data in the environment. To overcome these issues, the task of semantic occupancy prediction (SOP) is proposed, which aims to jointly infer both the geometry and semantic labels of a scene from images. However, conventional camera-based methods typically treat all categories equally and primarily rely on local features, leading to suboptimal predictions, especially for dynamic foreground objects. To address this, we propose Object-Centric SOP (OC-SOP), a framework that integrates high-level object-centric cues extracted via a detection branch into the semantic occupancy prediction pipeline. This object-centric integration significantly enhances the prediction accuracy for foreground objects and achieves state-of-the-art performance among all categories on SemanticKITTI.
\end{abstract}

\section{Introduction}
\label{sec:Introduction}
In current commercial autonomous driving frameworks, vision-only solutions primarily rely on multiple cameras mounted around the vehicle to perceive the surrounding environment. This approach not only mimics the way human drivers depend on vision but also significantly reduces system costs by eliminating expensive sensors like LiDAR. Such cost efficiency, combined with the rich contextual and high-resolution data captured by cameras, offers a promising pathway toward human-level driving capabilities and smart systems in a dynamic world. However, vision-only approaches face significant challenges in 3D scene understanding. Although state-of-the-art 2D semantic segmentation models can extract detailed pixel-wise semantic information from images, the absence of explicit depth information makes it difficult to accurately reconstruct 3D geometry and effectively map semantics into 3D space. Moreover, occlusions and perspective distortions inherent in single-view imaging result in large portions of the scene being partially invisible. This gap becomes especially challenging when trying to predict the structure and behavior of dynamic foreground objects, such as vehicles, pedestrians, and cyclists, which are critical for safe navigation in dynamic driving environments.

Monocular 3D Semantic Occupancy Prediction (SOP)~\cite{li2020anisotropic,cao2022monoscene,zhang2023occformer,huang2023tri} aims to infer a 3D semantic occupancy grid from a 2D image, effectively representing the geometry and semantics of the environment around the vehicle. Each spatial location in the grid records its occupancy and semantic information, enabling comprehensive environmental perception. However, the inherently ill-posed nature of this task poses significant challenges in diverse driving scenarios. Despite these difficulties, SOP has attracted significant attention for its potential to enhance the 3D perception capabilities of autonomous vehicles.

Traditional semantic occupancy prediction methods largely extend the 2D semantic segmentation paradigm into the 3D domain. Approaches such as U-Net–based hierarchical feature extraction or Transformer-based self-attention, which capture voxel-to-voxel dependencies, primarily rely on local feature extraction and contextual reasoning. Traditional methods treat each category equally by merely adjusting category weights based on voxel counts without additional design, resulting in significant limitations. In autonomous driving scenarios, background categories (e.g., roads and buildings) typically yield better prediction results due to their simple shapes, consistent textures, and spatial continuity. In contrast, foreground objects (e.g., vehicles, pedestrians, and cyclists) are much more challenging to predict because they exhibit complex geometries and diverse appearances. Moreover, since these foreground objects actively participate in driving scenarios and display far more intricate behaviors and interactions, accurately predicting them is critical for safe driving.

\begin{figure}[t]
    \centering
    \includegraphics[width=0.48\textwidth]{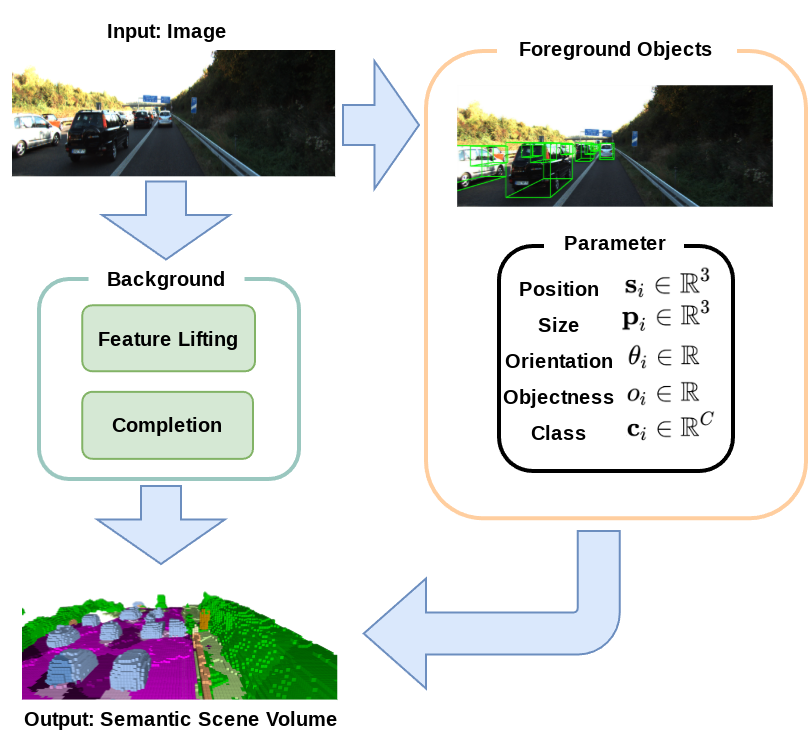}
    \caption{OC-SOP predicts a semantic occupancy volume from a single image using a 2D-to-3D feature lifting and completion pipeline, while enhancing foreground object-centric awareness via explicit extraction of bounding box parameters. Parts of the output scene lie outside of the field of view (FoV), which are visualized as shadow areas.}
    \label{fig:teaser}
    \vspace{-1.5em}
\end{figure}

To address these challenges, we propose Object-Centric SOP (OC-SOP), a framework that incorporates high-level object-centric awareness into the SOP pipeline. As shown in Fig.~\ref{fig:teaser}, object parameters such as pose and size provide robust constraints on object perception, ensuring that the predicted boundaries more closely align with the actual object dimensions. As a result, OC-SOP significantly improves object prediction accuracy, reduces the misclassification of empty space as part of adjacent objects, and mitigates distortions caused by camera effects. These enhancements ultimately boost the system’s overall environmental understanding and decision-making capabilities.

\noindent Our main contributions in this paper are:
\begin{itemize} 
	\item We introduce OC-SOP, a novel framework for 3D semantic occupancy prediction (SOP) that leverages high-level object-centric awareness by extracting object-centric cues using a detection-based branch and integrating these cues into the main completion branch.
	\item We propose a novel transformer module that tokenizes bounding box parameters into queries, which are then fused into the latent space of the completion U-Net.
	\item OC-SOP significantly improves the prediction accuracy for foreground objects and achieves state-of-the-art performance in camera-based semantic scene completion (SSC) on SemanticKITTI.
\end{itemize}

\section{Related Work}
\label{sec:Related Work}
\subsection{Camera-based 3D Reconstruction}
Camera-based 3D perception aligns with the natural human ability to form a holistic scene understanding from rich visual information. It is also increasingly favored in autonomous driving due to its cost-effectiveness. Early multi-view reconstruction~\cite{goesele2007multi, furukawa2010accurate, hu2025icg, hu20223d} and SLAM~\cite{engel2014lsd, mur2017orb} studies estimated 3D geometry from corresponding 2D feature points using explicit mathematical constraints. However, their reliance on multi-view observations restricted agent movement and limited scene reconstruction quality. Eigen et al.~\cite{eigen2014depth} first introduced end-to-end monocular depth estimation using a CNN, enabling 3D reconstruction from a single 2D image. This addressed the ill-posed problem of 2D-to-3D reconstruction. Inspired by~\cite{eigen2014depth}, some works focus on object-level single-view 3D reconstruction~\cite{wu2016learning, fan2017point, mescheder2019occupancy, peng2020convolutional}, typically employing an encoder-decoder structure to learn explicit or implicit geometry representations. Dahnert et al.~\cite{dahnert2021panoptic} expands this to scene-level reconstruction by lifting 2D panoptic features into 3D space. However, since single-view reconstruction focuses only on visible regions, it exhibits weak performance in predicting occluded areas, significantly hindering holistic 3D scene understanding. To address this, scene completion is proposed to predict unseen geometry based on observable information.

\subsection{Semantic Occupancy Prediction (SOP)}
Although there is some debate within the community regarding the definition of Semantic Occupancy Prediction (SOP), the widely accepted view is that SOP aims to predict occupancy status and semantics within a grid-based scene volume, covering both visible and occluded areas~\cite{su2024alpha}. Based on the input modality, SOP can be categorized into LiDAR-based SOP~\cite{roldao2020lmscnet, yan2021sparse, cao2025swasop}, Camera-based SOP~\cite{cao2022monoscene}, and Fusion-based SOP~\cite{cao2024slcf}, among others. A broader task is Semantic Scene Completion (SSC), which jointly predicts geometry and semantics for both visible and occluded regions~\cite{RoldaoCV:IJCV22, song2017semantic}. However, scene representation is not limited to grid-based formats; it also includes explicit representations like point clouds~\cite{cao2024diffssc} and meshes~\cite{dai2018scancomplete}, as well as implicit representations such as Neural Radiance Fields (NeRF)~\cite{nguyen2024semantically}. Voxel-based representations are an effective way to model 3D scenes, as they discretize space into voxels, each storing attributes like occupancy, semantics, or learned feature vectors. Due to their compatibility with 3D convolutional operations, early LiDAR-based SOP~\cite{roldao2020lmscnet, yan2021sparse} methods commonly used U-Net architectures to predict occluded regions. Later, the 3D completion U-Net was extended to camera-based approaches. MonoScene~\cite{cao2022monoscene} backprojects image features along optical rays to construct an initial voxel representation, which is then refined using a 3D completion U-Net. VoxFormer~\cite{li2023voxformer} introduced a Transformer-based approach, encoding spatial information as queries and context as key-value pairs, leveraging deformable attention for 3D-SOP. 

Most SOP methods focus on scene-level occupancy, treating background and foreground objects similarly. However, foreground objects exhibit fine-grained shapes and complex dynamics, making them more challenging for models to learn. Our work creates object-centric representations through an object detection branch, enabling the network to better capture shape details and refine object category predictions.

\subsection{Object-Centric Perception}
Human perception is fundamentally object-centric, as our cognitive system naturally focuses on distinct objects, which forms the basis for advanced cognition and effective interaction with the world. Recent works have exploited this principle in both 3D object detection~\cite{wang2025adaptive} and video prediction. For instance, inspired by the end-to-end paradigm of DETR~\cite{carion2020end}, DETR3D~\cite{wang2022detr3d} establishes learnable 3D object queries that link 2D images via camera projection matrices, while OCVP~\cite{villar2023object} decomposes video frames into object components and models their dynamics and interactions to generate future video frames. Leveraging object-centric awareness for semantic occupancy prediction is a novel direction. Unlike object detection—which outputs coarse bounding box representations—semantic occupancy prediction (SOP) produces finer, voxel-level semantic maps. Moreover, compared to video prediction, which typically focuses on visible regions, SOP must also predict occupancy in unseen or occluded areas, making the incorporation of object-centric awareness even more critical for accurate scene understanding.

\section{Method}
\label{sec:Methodology}
\begin{figure*}[!ht]
	%\vspace{-2.0em}
	\centering
	\includegraphics[width=\textwidth]{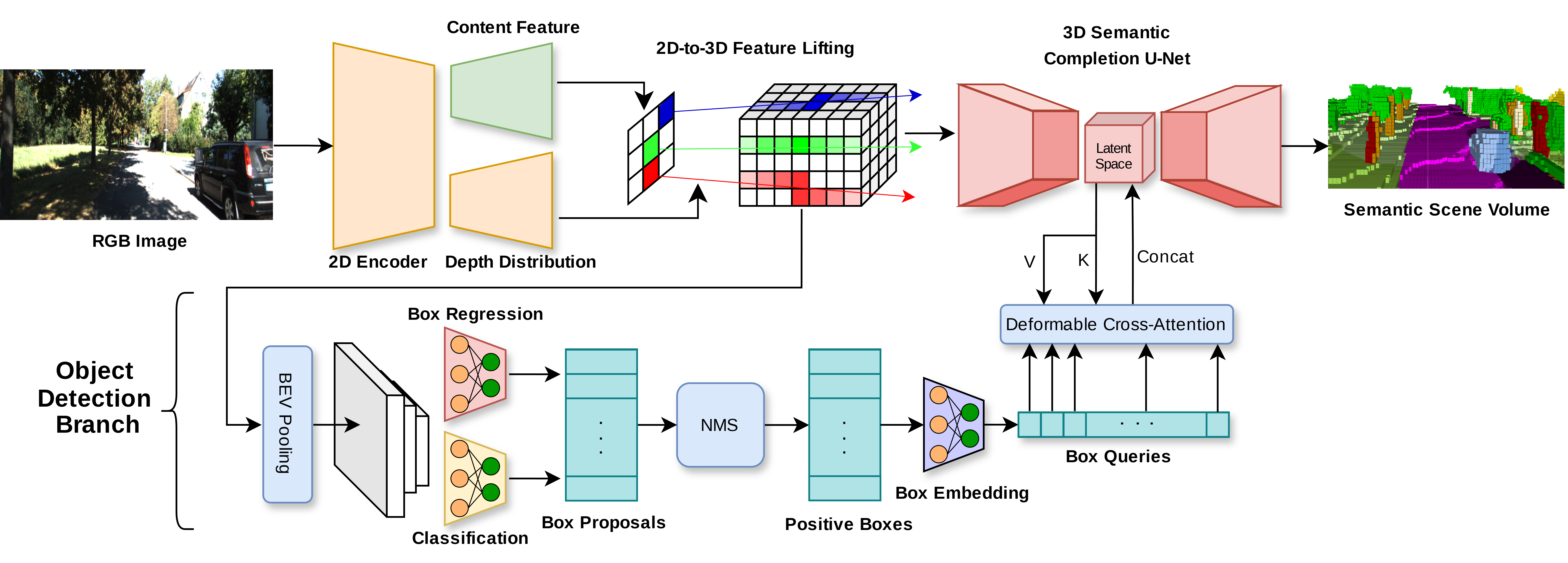}\vspace*{-.5em}
	\caption{The overall pipeline of OC-SOP can be divided into a main prediction branch (top) and an object detection-based branch (bottom). In the main branch, the RGB image is first fed into an encoder with dual decoders, where one decoder estimates depth and the other computes semantic content features. Next, a feature lifting module maps the content features into 3D space, and the resulting 3D features are processed by a 3D semantic completion U-Net that predicts the full semantic scene volume. In the object detection branch, the 3D semantic features are fed to a detection head to generate box proposals. Non-maximum suppression (NMS) is applied to select positive boxes. These positive boxes are then tokenized into queries via an MLP and fused into the completion network’s latent space through deformable cross-attention.}
	\label{fig:OC-SOP}
	\vspace*{-.5em}
\end{figure*}

We consider a monocular 3D Semantic Occupancy Prediction (SOP) task that jointly predict the occupancy status and semantic label of every voxel in a scene. In this task, a single RGB image $\mathbf{X}^{\text{RGB}}$ is used as input to produce a 3D semantic volume $\hat{\mathbf{Y}}$ defined over a target space $S$ with dimensions $(H, W, D)$. Every voxel in $\hat{\mathbf{Y}}$ is assigned one of $M+1$ labels from the set $\mathcal{C} = \{c_0, c_1, \dots, c_M\}$, where $c_0$ represents empty space and $c_1, \dots, c_M$ correspond to the semantic classes. 

As shown in Fig.~\ref{fig:OC-SOP}, our dual-branch network architecture~\cite{tao2023dudb} comprises a main prediction branch and an object detection-based branch, which share a common feature lifting backbone. In the backbone, we adopt an encoder with dual decoders (EDD), where one decoder estimates depth and the other computes semantic content features. The semantic content features, combined with the estimated depth, are then fed into a feature lifting module that maps the semantic features into 3D space.

The resulting 3D semantic features are subsequently processed by both branches. The main branch employs a 3D semantic completion U-Net to predict the full semantic scene, while the object detection branch utilizes an object detection head to generate box proposals for the individual objects in the scene. These object detections are later fused into the latent space of the main branch, thereby enhancing the object-centric understanding.

\subsection{Main Prediction Branch}
Our main prediction network is composed of a feature lifting backbone and a 3D semantic completion U-Net, which represents the classic paradigm for vision-based semantic occupancy prediction originally proposed by Cao and de Charette in MonoScene~\cite{cao2022monoscene}. MonoScene suffers from depth ambiguities caused by simply projecting 2D features along camera rays, which limits its ability to capture precise depth information. To address these issues, we enhance feature extraction by designing an encoder dual decoder (EDD).

Our encoder dual decoder (EDD) architecture disentangles semantic and geometric reasoning into two dedicated decoding branches. Starting from the shared 2D encoder feature map, one decoder branch extracts high-level semantic content, while the other estimates a dense per-pixel depth distribution over discretized depth bins. These two outputs are then combined through a depth-aware soft lifting process, where each 2D semantic feature is projected along its corresponding viewing ray, weighted by the predicted depth probabilities. This lifting operation is not limited to a single scale, but is performed hierarchically across multiple decoder layers, enabling the network to capture spatial correspondences at different levels of context.

Finally, the robust 3D features are fed into the 3D semantic completion U-Net, which predicts the complete semantic scene, including occluded regions, thereby significantly mitigating the camera effect. Our experiments reported in Sec.~\ref{sec:Experiment} show that even without fusing object-centric awareness, our approach achieves improved performance compared to MonoScene.

\subsection{Object Detection Branch}
We feed the 3D semantic features also to an object detection network to detect individual objects in the scene. The detection network predicts a set of box proposals 
$\mathbf{B} \in \mathbb{R}^{N \times D} = \{\mathbf{b}_0, \mathbf{b}_1, \dots, \mathbf{b}_N\}$, where each proposal $\mathbf{b}_i$ is a $D$-dimensional vector representing the box parameters, formulated as $\mathbf{b}_i = \left[\mathbf{p}_i, \mathbf{s}_i, \sin\theta, \cos\theta, \mathbf{c}_i, o_i\right]$. Here, $\mathbf{p}_i \in \mathbb{R}^3$ denotes the center position, $\mathbf{s}_i \in \mathbb{R}^3$ represents the size, $\theta \in \mathbb{R}$ is the orientation (expressed via its sine and cosine), $\mathbf{c}_i\in \mathbb{R}^C$ indicates the class, and $o_i\in \mathbb{R}$ is the objectness score.

Inspired by center-based 3D object detection such as VoteNet~\cite{qi2019deep} and PointRCNN~\cite{shi2019pointrcnn}, we adopt a lightweight detection head operating on lifted voxel features. The input to the detection module is a 3D feature volume $\mathbf{F}^{3D} \in \mathbb{R}^{C' \times D' \times H' \times W'} $, where $D'$ is the vertical dimension, and $H'$ and $W'$ correspond to the front-back and left-right axes in the bird’s eye view (BEV), respectively.

We first perform average pooling along the vertical axis to obtain a BEV representation $\mathbf{F}^{\text{BEV}} \in \mathbb{R}^{C' \times H' \times W'}$, which is then processed by a shallow 2D CNN for spatial feature extraction. From the resulting BEV feature map, we generate a dense objectness heatmap $\mathbf{H} \in \mathbb{R}^{1 \times H' \times W'}$, indicating the likelihood of object centers at each spatial location. Based on the top-K peaks in this heatmap, we select candidate positions and apply two independent 2-layer MLPs at those locations: one for regressing the 3D bounding box parameters (center, size, and orientation), and the other for predicting objectness scores and semantic class probabilities.

For each candidate predicted from the heatmap, we assign a positive label if its center is within 1\,m of any ground-truth object center, and a negative label if it is farther than 2\,m, ensuring that proposals with different labels contribute appropriately to the respective loss computations.

The detection loss is given by:
\begin{align}
L_\textrm{det} = L_\textrm{obj} + L_\textrm{reg} + L_\textrm{cls},
\end{align}
where the objectness loss $L_\textrm{obj}$ uses cross-entropy to optimize the model’s ability to distinguish between positive and negative samples. Negative proposals, which do not have a corresponding ground truth box, are only used to compute the objectness loss, while their box parameters and class components are ignored. Proposals that are neither clearly positive nor negative do not contribute to the loss. For positive proposals, we compute both a regression loss $L_\textrm{reg}$ and a semantic classification loss $L_\textrm{cls}$. The regression loss, implemented using Smooth L1, measures errors in the predicted box parameters (center position, size, and orientation), with different weights assigned to each component to balance their contributions. The semantic classification loss uses multi-class cross-entropy to ensure accurate category prediction. 

\subsection{Box Feature Fusion}
After obtaining a set of box proposals, we apply non-maximum suppression (NMS) to select the final box features. In our approach, each selected box proposal is first encoded into a query vector using an MLP. For a given query $q$, we define $K$ sampling points in the latent space, where each sampling location is determined by a base position $p$ plus a learned offset $\Delta p_k$ (for $k = 1, \dots, K$). The deformable attention mechanism then computes the output as follows:
\begin{align}
\mathrm{Output}(q)= \sum_{k=1}^{K} A_{k}\, V\bigl(p+\Delta p_{k}\bigr),
\end{align}
where $A_{k}$ denotes the attention weight at the $k$th sampling point and $V(\cdot)$ is the value function that extracts the corresponding feature from the latent space. This output is then fused into the latent space of the 3D semantic completion U-Net (via additional convolutional layers or residual fusion), thereby integrating complete query information with adaptively sampled latent features for effective fusion.

\subsection{Progressive Multi-task Training Strategy}
\begin{figure}[t]
    \centering
    \includegraphics[width=0.48\textwidth]{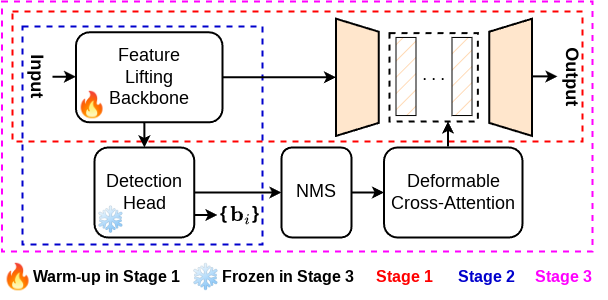}
    \caption{Three-stage training strategy of OC-SOP.}
    \label{fig:train}
    \vspace{-1.em}
\end{figure}

Considering that our framework exhibits multi-task characteristics, in which the feature lifting backbone is utilized by both the main and detection branches, we adopt a three-stage training strategy, as shown in Fig.~\ref{fig:train}. 

In Stage~1, we disable the detection branch and train only the feature lifting backbone along with the 3D completion U-Net. The goal in this phase is to warm up the backbone so that it efficiently extracts depth and semantic content information from the images and forms robust 3D features. 

In Stage~2, we remove the 3D completion U-Net and load the pre-trained backbone to integrate the detection head module, with both positive and negative boxes used for loss computation. 

In the final Stage~3, we freeze the detection head and apply non-maximum suppression (NMS) to select high-confidence, object-centric features. At this point, the overall network performance is further improved by effectively integrating the stable detection outputs into the fusion process. 

This three-stage strategy is highly effective in multi-task learning scenarios for stabilizing training and achieving the reported results.

\section{Experiment}
\label{sec:Experiment}
\definecolor{carColor}{RGB}{100,150,245}
\definecolor{bicycleColor}{RGB}{100,230,245}
\definecolor{motorcycleColor}{RGB}{30,60,150}
\definecolor{truckColor}{RGB}{80,30,180}
\definecolor{othervehicleColor}{RGB}{0,0,255}
\definecolor{personColor}{RGB}{255,30,30}
\definecolor{bicyclistColor}{RGB}{255,40,200}
\definecolor{motorcyclistColor}{RGB}{150,30,90}
\definecolor{roadColor}{RGB}{255,0,255}
\definecolor{parkingColor}{RGB}{255,150,255}
\definecolor{sidewalkColor}{RGB}{75,0,75}
\definecolor{othergroundColor}{RGB}{175,0,75}
\definecolor{buildingColor}{RGB}{255,200,0}
\definecolor{fenceColor}{RGB}{255,120,50}
\definecolor{vegetationColor}{RGB}{0,175,0}
\definecolor{trunkColor}{RGB}{135,60,0}
\definecolor{terrainColor}{RGB}{150,240,80}
\definecolor{poleColor}{RGB}{255,240,150}
\definecolor{trafficsignColor}{RGB}{255,0,0}

\begin{table*}[t]
%\vspace{-.5em}
\caption{Quantitative results on the SemanticKITTI test set.}
\vspace{-.5em}
\label{tab:evaluation}
\centering
\renewcommand{\arraystretch}{1.1} %<- modify value to suit your needs
\linespread{0.95}\selectfont
\setlength{\tabcolsep}{1.3pt}
\sisetup{separate-uncertainty}
%\resizebox{\columnwidth}{!}{%
\begin{threeparttable}
\begin{tabular}{c|c|c|cccccccc|ccccccccccc}%L{2.3cm}
  \toprule
  & & &\multicolumn{8}{c|}{Objects} & \multicolumn{11}{c}{Background}\\
   Method & IoU & mIoU & 
	\rotatebox{90}{\tikz{\fill[carColor] (0,0) rectangle (0.2cm,0.2cm);} \textbf{car} \tiny{(3.92$\%$)}} &
	\rotatebox{90}{\tikz{\fill[truckColor] (0,0) rectangle (0.2cm,0.2cm);} \textbf{truck} \tiny{(0.16$\%$)}} &
	\rotatebox{90}{\tikz{\fill[bicycleColor] (0,0) rectangle (0.2cm,0.2cm);} \textbf{bicycle} \tiny{(0.03$\%$)}} &
	\rotatebox{90}{\tikz{\fill[motorcycleColor] (0,0) rectangle (0.2cm,0.2cm);} \textbf{motorcycle} \tiny{(0.03$\%$)}} &
	\rotatebox{90}{\tikz{\fill[othervehicleColor] (0,0) rectangle (0.2cm,0.2cm);} \textbf{other-vehicle} \tiny{(0.20$\%$)}} &
	\rotatebox{90}{\tikz{\fill[personColor] (0,0) rectangle (0.2cm,0.2cm);} \textbf{person} \tiny{(0.07$\%$)}} &
	\rotatebox{90}{\tikz{\fill[bicyclistColor] (0,0) rectangle (0.2cm,0.2cm);} \textbf{bicyclist} \tiny{(0.07$\%$)}} &
	\rotatebox{90}{\tikz{\fill[motorcyclistColor] (0,0) rectangle (0.2cm,0.2cm);} \textbf{motorcyclist} \tiny{(0.05$\%$)}} &
	\rotatebox{90}{\tikz{\fill[roadColor] (0,0) rectangle (0.2cm,0.2cm);} \textbf{road} \tiny{(15.30$\%$)}} &
	\rotatebox{90}{\tikz{\fill[parkingColor] (0,0) rectangle (0.2cm,0.2cm);} \textbf{parking} \tiny{(1.12$\%$)}} &
	\rotatebox{90}{\tikz{\fill[sidewalkColor] (0,0) rectangle (0.2cm,0.2cm);} \textbf{sidewalk} \tiny{(11.13$\%$)}} &
	\rotatebox{90}{\tikz{\fill[othergroundColor] (0,0) rectangle (0.2cm,0.2cm);} \textbf{other-ground} \tiny{(0.56$\%$)}} &
	\rotatebox{90}{\tikz{\fill[buildingColor] (0,0) rectangle (0.2cm,0.2cm);} \textbf{building} \tiny{(14.10$\%$)}} &
	\rotatebox{90}{\tikz{\fill[fenceColor] (0,0) rectangle (0.2cm,0.2cm);} \textbf{fence} \tiny{(3.90$\%$)}} &
	\rotatebox{90}{\tikz{\fill[vegetationColor] (0,0) rectangle (0.2cm,0.2cm);} \textbf{vegetation} \tiny{(39.30$\%$)}} &
	\rotatebox{90}{\tikz{\fill[trunkColor] (0,0) rectangle (0.2cm,0.2cm);} \textbf{trunk} \tiny{(0.51$\%$)}} &
	\rotatebox{90}{\tikz{\fill[terrainColor] (0,0) rectangle (0.2cm,0.2cm);} \textbf{terrain} \tiny{(9.17$\%$)}} &
	\rotatebox{90}{\tikz{\fill[poleColor] (0,0) rectangle (0.2cm,0.2cm);} \textbf{pole} \tiny{(0.29$\%$)}} &
	\rotatebox{90}{\tikz{\fill[trafficsignColor] (0,0) rectangle (0.2cm,0.2cm);} \textbf{traffic-sign} \tiny{(0.08$\%$)}} \\
  \midrule
   LMSCNet*~\cite{roldao2020lmscnet} &31.38  & 7.07 &
   14.30 & 0.30 & 0.00 & 0.00 & 0.00 & 
   0.00 &0.00  & 0.00 & 46.70 & 13.50 &
   19.50 & 3.10 & 10.30 &  5.40& 10.80 &
   0.00 & 10.40 & 0.00 & 0.00 \\
   
   AICNet*~\cite{li2020anisotropic} & 23.93 & 7.09 &
   15.30 & 0.70 & 0.00 & 0.00 & 0.00 & 
   0.00 & 0.00 & 0.00 &  
   39.30 & 19.80 & 18.30 & 1.60 & 9.60 &  
   5.00 & 9.60 & 1.90 & 13.50 & 0.10 & 0.00 \\
   
   JS3C-Net*~\cite{yan2021sparse} & 34.00 & 8.97 &
   20.10 & 0.80 & 0.00 & 0.00 & 4.10 & 
   0.00 & 0.20 & 0.20 &  
   47.30 & 19.90 & 21.70 & 2.80 & 12.70 & 
   8.70 & 14.20 & 3.10 & 12.40 & 1.90 & 0.30 \\
    
   MonoScene~\cite{cao2022monoscene} & 34.16 & 11.08 &
   18.80 & 3.30 & 0.50 & 0.70 & \textbf{4.40} & 
   1.00 & 1.40 & \underline{0.40} & 
   54.70 & 24.80 & 27.10 & 5.70 & 14.40 & 
   11.10 & 14.90 & 2.40 & 19.50 & 3.30 & 2.10 \\
    
   TVPFormer~\cite{huang2023tri} & 34.25 & 11.26 &
   19.20 & \underline{3.70} & 1.00 & 0.50 & 2.30 &
   1.10 & 2.40 & 0.30 & 
   55.10 & \textbf{27.40} & 27.20 & \underline{6.50} & 14.80 &
   11.00 & 13.90 & 2.60 & 20.40 & 2.90 & 1.50 \\
    
   VoxFormer~\cite{li2023voxformer} & \underline{42.95} & 12.20 &
   20.80 & 3.50 & 1.00 & 0.70 & \underline{3.70} & 
   1.40 & \underline{2.60} & 0.20 &
   53.90 & 21.10 & 25.30 & 5.60 & \underline{19.80} &
   11.10 & \textbf{22.40} & \underline{7.50} & \underline{21.30} & \underline{5.10} & \underline{4.90}\\
    
   OccFormer~\cite{zhang2023occformer} & 34.53 & \underline{12.32} &
   \underline{21.60} & 1.20 & \underline{1.50} & \underline{1.70} & 3.20 & 
   \underline{2.20} & 1.10 & 0.20 &  
   \underline{55.90} & 31.50 & \textbf{30.30} & 6.50 & 15.70 &
   \underline{11.90} & 16.80 & 3.90 & \underline{21.30} & 3.80 & 3.70\\
   \midrule
    OC-SOP &\textbf{43.30}  &\textbf{14.83}  &
    \textbf{30.40} & \textbf{4.80}  &\textbf{7.80}  &\textbf{6.30}  &3.50  & 
    \textbf{5.80} & \textbf{3.70}  & \textbf{1.30}  &
    \textbf{56.00}  & \underline{27.10} & \underline{29.50} 
    & \textbf{7.10} & \textbf{21.50} & \textbf{13.30} & \underline{20.80} &
    \textbf{8.30} & \textbf{22.20} & \textbf{5.90} & \textbf{6.40} \\
   \bottomrule
\end{tabular}
\end{threeparttable}
%}

\vspace*{1ex} 

\footnotesize{All listed baselines are vision-based methods. * denotes the reproduced results reported by~\cite{cao2022monoscene}.\\IoU focuses solely on the occupancy status while mIoU evaluates individual semantic categories. \textbf{Best} and \underline{second best} results are highlighted.}
\vspace*{-2ex}
\end{table*}

\subsection{Experiment Setup}
\subsubsection{Datasets}
We train our model using two datasets. For voxel-wise prediction, we employ the SemanticKITTI~\cite{behley2019iccv} dataset, on which numerous classic baselines have been validated. Built on the KITTI odometry benchmark, SemanticKITTI enables to study semantic segmentation through manual annotations on raw LiDAR point clouds. The semantic scene completion benchmark is constructed by accumulating annotated scans within sequences and voxelizing them to form a semantic scene volume with dimensions of 256$\times$256$\times$32 (with each voxel representing 0.2\,m). The dataset contains 21 classes, including 19 semantic labels, one free label, and one unknown label. In our experiments, we use the RGB images from cam-2 with a resolution of 1226$\times$370 (cropped to 1220$\times$370) and follow the official training/validation split of 3834 and 815 samples, respectively.

For detection branch training, we utilize the KITTI~\cite{geiger2012cvpr} 3D object detection benchmark dataset, which comprises 7481 training images and 7518 test images along with their corresponding point clouds. We follow the train/validation split defined by Chen et al.~\cite{chen2016monocular}, dividing the training set into 3712 training images and 3769 validation images. KITTI annotates three foreground object categories (Car, Pedestrian, and Cyclist), which completely align with our definition of foreground objects and are well suited for object-centric tasks on SemanticKITTI. Although the KITTI's semantic labels are coarser than the fine-grained semantic labels in SemanticKITTI, these rough labels still provide a semantic prior for foreground objects in semantic occupancy prediction tasks.

\subsubsection{Implementation Details}
The model is trained on an NVIDIA A6000 GPU in three stages: a warm-up Stage~1 for 5 epochs, a detection branch training Stage~2 for 10 epochs, and a joint fine-tuning Stage~3 for 10 epochs. During joint fine-tuning and inference, we set the detection head’s objectness threshold to 0.2 to filter out low-score negative predictions, followed by applying an NMS module to obtain the final box proposals. The NMS IoU threshold is set to 0.7.

\subsection{Main Results}

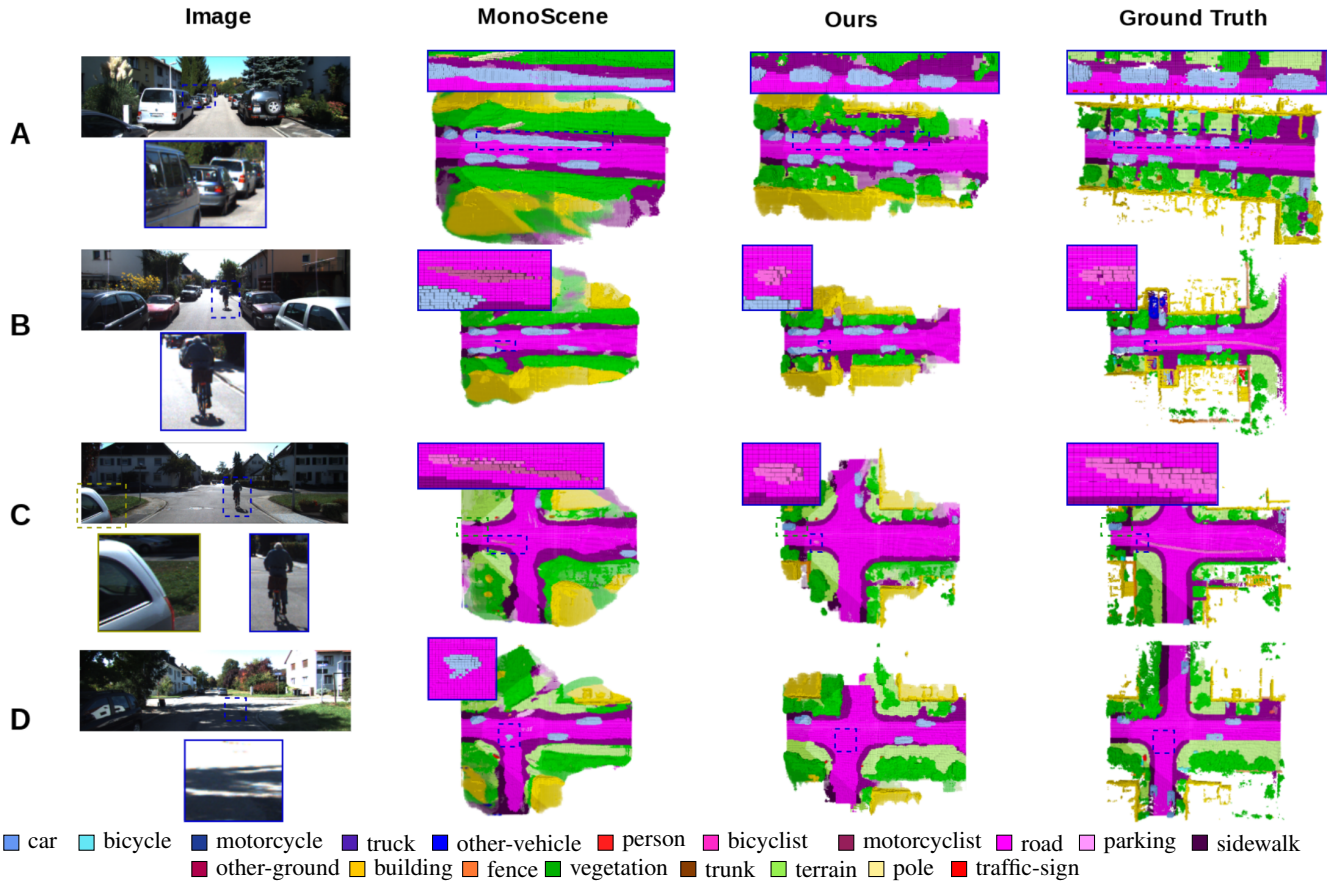
\begin{figure*}[t]
  \centering
  \begin{tikzpicture}
    \node[anchor=south west,inner sep=0] (image) at (0,0) {\includegraphics[width=\textwidth]{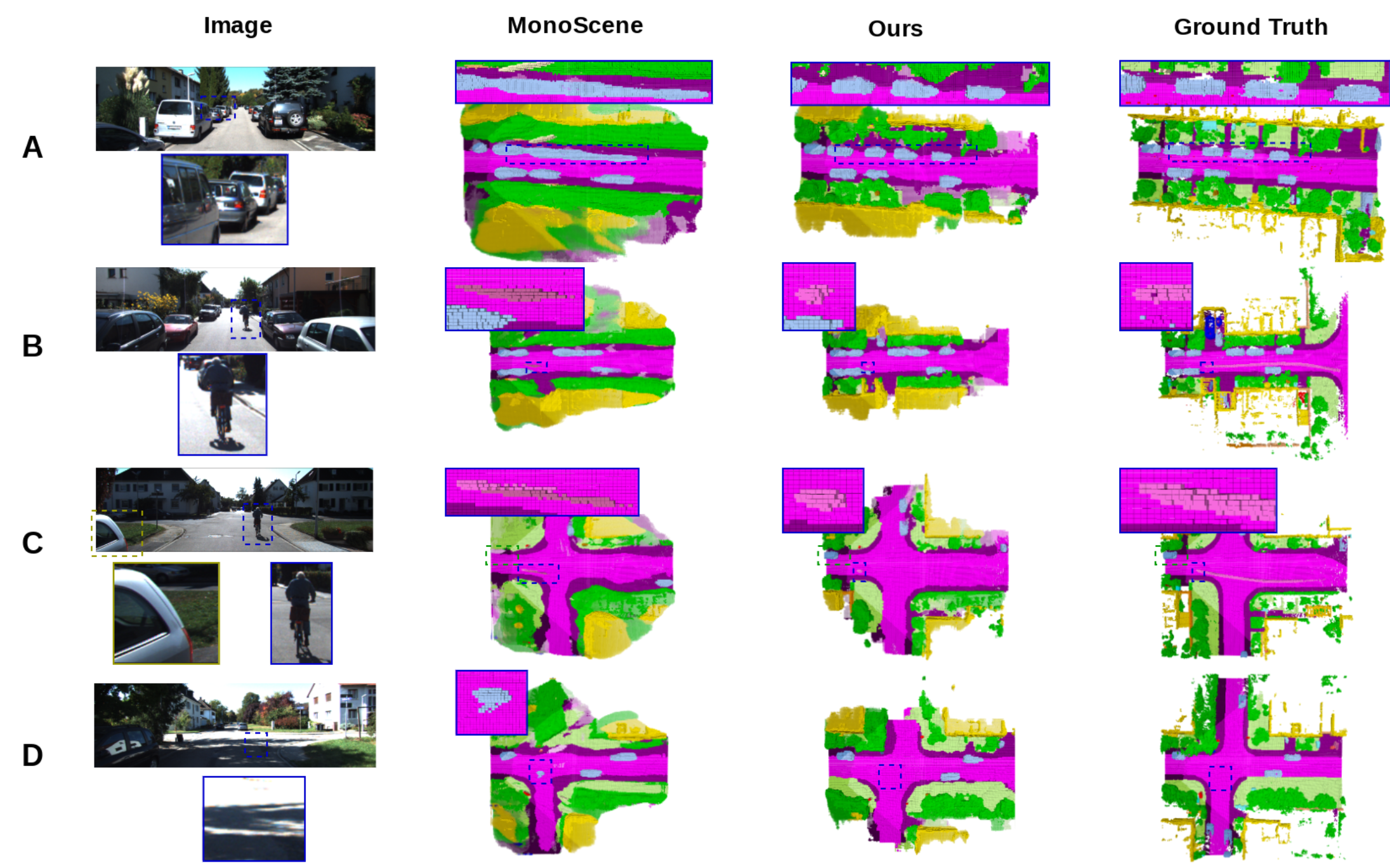}};
    \def\legendWidth{2.0}
    \def\legendHeight{0.2}
    \def\legendSpacing{0.15}
    \def\yOffset{-0.3}
    
   \draw[fill=carColor] ($ (image.south west) + (0.2, \yOffset)$) rectangle ++(0.2, 0.2) node[right, yshift=-0.1cm] {\small car};
	\draw[fill=bicycleColor] ($ (image.south west) + (1.2, \yOffset)$) rectangle ++(0.2, 0.2) node[right, yshift=-0.1cm] {\small bicycle};
    \draw[fill=motorcycleColor] ($ (image.south west) + (2.7, \yOffset)$) rectangle ++(0.2, 0.2) node[right, yshift=-0.1cm] {\small motorcycle};
    \draw[fill=truckColor] ($ (image.south west) + (4.7, \yOffset)$) rectangle ++(0.2, 0.2) node[right, yshift=-0.1cm] {\small truck};
    \draw[fill=othervehicleColor] ($ (image.south west) + (5.9, \yOffset)$) rectangle ++(0.2, 0.2) node[right, yshift=-0.1cm] {\small other-vehicle};
    \draw[fill=personColor] ($ (image.south west) + (8.1, \yOffset)$) rectangle ++(0.2, 0.2) node[right, yshift=-0.1cm] {\small person};
    \draw[fill=bicyclistColor] ($ (image.south west) + (9.5, \yOffset)$) rectangle ++(0.2, 0.2) node[right, yshift=-0.1cm] {\small bicyclist};
    \draw[fill=motorcyclistColor] ($ (image.south west) + (11.3, \yOffset)$) rectangle ++(0.2, 0.2) node[right, yshift=-0.1cm] {\small motorcyclist};
    \draw[fill=roadColor] ($ (image.south west) + (13.4, \yOffset)$) rectangle ++(0.2, 0.2) node[right, yshift=-0.1cm] {\small road};
    \draw[fill=parkingColor] ($ (image.south west) + (14.5, \yOffset)$) rectangle ++(0.2, 0.2) node[right, yshift=-0.1cm] {\small parking};
    \draw[fill=sidewalkColor] ($ (image.south west) + (16.0, \yOffset)$) rectangle ++(0.2, 0.2) node[right, yshift=-0.1cm] {\small sidewalk};
    
    \draw[fill=othergroundColor] ($ (image.south west) + (2.7, \yOffset-\legendHeight - \legendSpacing)$) rectangle ++(0.2, 0.2) node[right, yshift=-0.1cm] {\small other-ground};
    \draw[fill=buildingColor] ($ (image.south west) + (4.8, \yOffset-\legendHeight - \legendSpacing)$) rectangle ++(0.2, 0.2) node[right, yshift=-0.1cm] {\small building};
        \draw[fill=fenceColor] ($ (image.south west) + (6.3, \yOffset-\legendHeight - \legendSpacing)$) rectangle ++(0.2, 0.2) node[right, yshift=-0.1cm] {\small fence};
    \draw[fill=vegetationColor] ($ (image.south west) + (7.4, \yOffset-\legendHeight - \legendSpacing)$) rectangle ++(0.2, 0.2) node[right, yshift=-0.1cm] {\small vegetation};
    \draw[fill=trunkColor] ($ (image.south west) + (9.2, \yOffset-\legendHeight - \legendSpacing)$) rectangle ++(0.2, 0.2) node[right, yshift=-0.1cm] {\small trunk};
    \draw[fill=terrainColor] ($ (image.south west) + (10.4, \yOffset-\legendHeight - \legendSpacing)$) rectangle ++(0.2, 0.2) node[right, yshift=-0.1cm] {\small terrain};
    \draw[fill=poleColor] ($ (image.south west) + (11.7, \yOffset-\legendHeight - \legendSpacing)$) rectangle ++(0.2, 0.2) node[right, yshift=-0.1cm] {\small pole};
    \draw[fill=trafficsignColor] ($ (image.south west) + (12.8, \yOffset-\legendHeight - \legendSpacing)$) rectangle ++(0.2, 0.2) node[right, yshift=-0.1cm] {\small traffic-sign};
  \end{tikzpicture}
	
	\vspace*{-1ex}

  \caption{Qualitative results on the SemanticKITTI validation set. Only the 19 semantic classes are visualized; voxels labeled as empty space are omitted for clarity. Voxels located in unknown regions are rendered with $20\%$ opacity to reflect uncertainty. Parts of the scene that lie outside the field of view (FoV) are visualized as shaded areas.}
  \label{fig:visualization}
  \vspace{-1.em}
\end{figure*}

We evaluate our model on the SemanticKITTI test set using two metrics. Intersection over Union (IoU) metrics focus only on occupancy status without considering semantic labels. Mean IoU (mIoU) is used to evaluate the accuracy of the semantic labels. Specifically, the IoU for each semantic category is first computed, and then the average of these values is taken to reflect overall performance. All IoU calculations exclude the unknown category. The results are reported in Tab.~\ref{tab:evaluation}, where we also compare our approach with several vision-based baselines on the SemanticKITTI benchmark. The results show that OC-SOP significantly outperforms all baselines on foreground object categories. For instance, when considering only foreground object categories, OC-SOP achieves an mIoU$_\textrm{obj}$ of 7.95\%, which is 3.71 percentage points higher than the second-best method, VoxFormer, which achieves an mIoU$_\textrm{obj}$ of 4.24\%. Moreover, even though background categories do not directly benefit from object-centric awareness, the design of our EDD and 3D semantic completion U-Net still achieves comparable performance, slightly outperforming most baselines.

Fig.~\ref{fig:visualization} shows several qualitative results from the SemanticKITTI validation set. We compare our predictions with those of MonoScene to highlight the benefits of incorporating object-centric awareness. In Scene~A, multiple vehicles are parked along both sides of the road with narrow gaps between them. MonoScene merges these vehicles into one block, making it difficult to identify the number of vehicles and resulting in severely distorted shapes. In contrast, our method clearly distinguishes each vehicle, preserving distinct gaps and maintaining shape integrity. In Scene~B, a bicyclist appears in front of the ego-vehicle. MonoScene misclassifies the bicyclist as a motorcyclist, and its shape is significantly distorted by camera effects. Our predictions, guided by bounding box constraints, more accurately capture the bicyclist's shape. It is important to note that since SemanticKITTI's ground truth is derived from aggregating multiple frames, dynamic objects may exhibit ghosting artifacts, a limitation that partly contributes to the misclassifications observed in MonoScene. Scene~C presents a similar issue with a bicyclist, and in addition, MonoScene fails to detect vehicles near the field-of-view boundaries. This suggests that traditional SOP methods are less sensitive to objects that are severely truncated in the view frustum, even though such objects, often located near the ego-vehicle, are vital for accurate scene understanding. Finally, in Scene~D, MonoScene mistakenly predicts a vehicle in an area where none exists, likely due to misleading lighting and shadow cues. The error from the semantic segmentation module is not corrected by the reprojection process and thus appears in the final output. In practical applications, this could lead to incorrect estimation of dynamic variables such as speed and trajectory, thereby affecting the comfort and even the safety of passengers and other traffic participants. In our framework, however, the objectness score helps to correct such misestimations, preventing these hallucinations in the final prediction.

\subsection{Ablation Studies}
\begin{table}[h]
%\vspace{-.5em}
\caption{Ablation study of each sub-module on the SemanticKITTI validation set.}
\vspace{-.5em}
\label{tab:ablation}
\centering
\begin{threeparttable}
\begin{tabular}{cc|c|c|c|c}%L{2.3cm}
  \toprule
  \multicolumn{2}{c|}{\textbf{Setting}}  & I & II & III & IV \\
  \midrule
	\multicolumn{2}{c|}{Dual Decoder}  & \checkmark &\checkmark &\checkmark & -\\
	\multicolumn{2}{c|}{Detection Branch} &\checkmark & \checkmark &- &-\\
	\multicolumn{2}{c|}{Deformable Cross-attention} & \checkmark &- &- &- \\
	\midrule
	\multirow{8}{*}{\rotatebox{90}{\textbf{Objects}}} & car   & \textbf{30.40} &  \underline{25.60} & 17.80 & 17.30\\
	& truck         &  \textbf{4.80}  &   \underline{3.90}    &   3.10    & 0.70\\
	& bicycle       &  \textbf{7.80}  &    \underline{5.30}   &   0.40    & 0.00\\
	& motorcycle    &  \textbf{6.30}  &    \underline{4.20}   &   0.70    & 0.00\\
	& other-vehicle &  \textbf{3.50}  &    \underline{3.40}   &   2.50   & 0.00\\
	& person        &  \textbf{5.80}  &    \underline{4.40}   &   1.50    & 0.00\\
	& bicyclist     &  \textbf{3.70}  &    \underline{3.10}   &   1.10    & 0.00\\
	& motorcyclist  &  \textbf{1.30}  &    \underline{1.20}   &   0.20    & 0.00\\
	\midrule
	\multirow{11}{*}{\rotatebox{90}{\textbf{Background}}} & road    & \textbf{56.00}  &  \underline{55.80}     & 55.30    & 46.30 \\
	& parking      &  \underline{27.10}    &   26.50    &   \textbf{27.70}    & 12.10\\
	& sidewalk     &  \textbf{29.50}     &   \underline{29.10}    &   28.20    & 20.10\\
	& other-ground &  \underline{7.10}     &   6.90    &    \textbf{7.30}   & 4.20\\
	& building     &  \textbf{21.50}     &    \underline{19.90}   &   18.70    & 10.70\\
	& fence        &  \textbf{13.30}     &   \underline{12.90}    &   10.90    & 6.90\\
	& vegetation   &  \textbf{20.80}     &    \underline{20.50}   &    19.00   & 10.30\\
	& trunk        &  \underline{8.30}     &    \textbf{8.50}   &    6.30   & 1.30\\
	& terrain      &  \underline{22.20}     &   21.50    &    \textbf{22.90}   & 11.20\\
	& pole         &  \textbf{5.90}     &    \underline{5.50}   &    4.80   & 0.30\\
	& traffic-sign &  \textbf{6.40}     &   \underline{6.10}    &    5.80   & 0.10\\
	\midrule
	\multicolumn{2}{c|}{IoU}  &  \textbf{43.30}     &    \underline{41.98}   &    41.52   & 29.78\\
	\multicolumn{2}{c|}{mIoU} &  \textbf{14.56}     &   \underline{13.92}    &   12.33    & 7.45\\
   \bottomrule
\end{tabular}
\end{threeparttable}

\vspace*{1ex} 
\footnotesize{IoU focuses solely on the occupancy status while mIoU evaluates indivi-\\dual semantic categories. \textbf{Best} and \underline{second best} results are highlighted.}
\vspace*{-2ex}

\end{table}
We conducted ablation studies on the SemanticKITTI validation set to assess the contributions of each sub-module.

In Setting~I, we employ the full model, which achieves the best performance for all foreground object categories and most background categories, yielding overall best results. 

In Setting~II, to isolate the impact of the box feature fusion module, we replace the deformable cross attention with spatial concatenation fusion. Specifically, each box proposal’s feature is first transformed using an MLP and then concatenated along the channel dimension with the main branch’s latent space. Subsequent convolution layers fuse these features. The results indicate that although Setting~II still incorporates object-centric awareness, the deformable cross attention used in Setting~I more effectively integrates box proposals with the latent features. This leads to a notable improvement in foreground object performance. For background categories, the performance of Setting~II is slightly lower than that of Setting~I, but the difference is not as pronounced as for foreground objects.

In Setting~III, only the main prediction branch is used for inference without any object-centric awareness. Comparing Setting~I with Setting~III, we observe that incorporating object-centric awareness significantly enhances the estimation of foreground objects, with performance improving by approximately 4.54 percentage points (from 3.41\% to 7.95\%) while background categories show only a modest improvement of around 1.02 percentage points (from 18.81\% to 19.83\%). Notably, the main prediction branch consists of an encoder dual decoder (EDD) and a 3D completion U-Net. This architecture effectively enhances the extraction of depth information, thereby mitigating camera effects, and still achieves competitive performance that slightly outperforms MonoScene.

Finally, Setting~IV retains only the 3D semantic completion U-Net, which relies solely on the raw image and an external depth predictor to form a pseudo-LiDAR 3D map. Due to the lack of a robust feature extractor, Setting~IV shows poor performance, with limited ability to predict several foreground categories and weak performance on background categories. The comparison between Settings~III and IV clearly demonstrates the effectiveness of the EDD in enhancing feature extraction for semantic occupancy prediction.

\section{Conclusions and Outlook}
\label{sec:Conclusion and Outlook}
We propose OC-SOP, which innovatively integrates object-centric awareness into the semantic occupancy prediction task. Our results on the SemanticKITTI test set demonstrate that OC-SOP significantly improves the accuracy of foreground object predictions. By focusing on other traffic participants, our approach enhances the perception system's ability to understand complex traffic scenarios, ultimately contributing to safer decision-making in autonomous driving. In future work, we plan to explore emerging techniques in prompt-driven representation learning~\cite{zhou2024source, li2025self, wu2025prompt} and multimodal fusion~\cite{wu2025llm, wuimgfu} to further improve object-centric semantics and generalization across diverse sensing conditions.

%\IEEEtriggeratref{20}
%when using cite package:
\bibliographystyle{IEEEtran}
\bibliography{literature}

\end{document}